\documentclass[runningheads]{llncs}
\usepackage[T1]{fontenc}
\usepackage{graphicx}
\usepackage{pdfpages}
\usepackage{comment}

\usepackage{todonotes}
\usepackage{subcaption}

\usepackage{hyperref}
\hypersetup{
    colorlinks=true,
    breaklinks=true,
    urlcolor= blue,
    linkcolor= blue,
    bookmarksopen=false,
    filecolor=black,
    citecolor=blue,
    linkbordercolor=blue
}

\usepackage{ amssymb }
\usepackage[misc,geometry]{ifsym}
\usepackage{orcidlink}

\usepackage{tabularray}
\usepackage{booktabs}
\usepackage{arydshln}
\usepackage{lipsum}
\usepackage{color}
\definecolor{Seashell}{rgb}{0.945,0.945,0.945}
\usepackage{mdframed}
\usepackage{array}

\usepackage{rotating}

\usepackage{url}

\usepackage{draftwatermark}

\SetWatermarkText{Preprint} 
\SetWatermarkScale{1} 
\SetWatermarkColor[gray]{0.9} 
\SetWatermarkAngle{45} 

\begin{document}
%

\title{Multi-Faceted Evaluation of Modeling Languages for Augmented Reality Applications - \\The Case of ARWFML}

\titlerunning{Multi-Faceted Evaluation of Modeling Languages for AR}
%
\author{Fabian Muff \inst{1} (\Letter)\orcidlink{0000-0002-7283-6603} \and Hans-Georg Fill\inst{1}\orcidlink{0000-0001-5076-5341}}
\authorrunning{Muff and Fill}
%
\institute{University of Fribourg, Research Group Digitalization and Information Systems\thanks{Financial support is gratefully acknowledged by the \href{https://www.smartlivinglab.ch/en/}{Smart Living Lab} funded by the University of Fribourg, EPFL, and HEIA-FR.}
\email{fabian.muff|hans-georg.fill@unifr.ch}
}

\maketitle            
\begin{abstract} The evaluation of modeling languages for augmented reality applications poses particular challenges due to the three-dimensional environment they target. The previously introduced Augmented Reality Workflow Modeling Language (ARWFML) enables the model-based creation of augmented reality scenarios without programming knowledge. Building upon the first design cycle of the language's specification, this paper presents two further design iterations for refining the language based on multi-faceted evaluations. These include a comparative evaluation of implementation options and workflow capabilities, the introduction of a 3D notation, and the development of a new 3D modeling environment. On this basis, a comprehensibility study of the language was conducted. Thereby, we show how modeling languages for augmented reality can be evolved towards a maturity level suitable for empirical evaluations.

\end{abstract}

\section{Introduction}
\label{sec:introduction}

Augmented Reality (AR) software applications are widely used today to superimpose any type of virtual information on the real world to enhance human perception, e.g., for purposes such as machine maintenance, training, or medical applications~\cite{xue2019virtual}. In the past, several model-based approaches have been proposed to facilitate the creation of such applications, e.g.,~\cite{lenk2012model,ruminski2014dynamic,grambow2021context,lechner2013arml,ruiz_rube2020model,campos_l_opez2023model,wild2020ieee}. At their core, they aim to simplify the creation of AR scenarios by freeing users from the need for specialized programming skills. Thereby, these modeling languages have to satisfy a number of specific requirements to cover basic and advanced augmented reality concepts. In addition, the inherent focus on three-dimensional environments poses additional challenges for the design and usage of such languages. In a previous design cycle, the Augmented Reality Workflow Modeling Language (ARWFML) was proposed as a novel approach to modeling AR scenarios. As a result of the first design cycle, the approach was demonstrated to encompass a more extensive range of AR concepts than any preceding approach, thereby conferring to a greater degree of expressiveness~\cite{muff2023domain}. In the course of a second design cycle, the language was implemented on the two-dimensional (2D) \textit{ADOxx} metamodeling platform. The goal of this implementation was to assess the technical feasibility of the language specification by translating it into the \textit{ADOxx} metamodeling concepts and to provide the basis for subsequent practical evaluations. Although this implementation was successfully achieved, some shortcomings were identified due to the limitation of \textit{ADOxx} to 2D modeling environments~\cite{fill2013conceptualisation}.

\begin{figure}[t!]
    \centering
    \includegraphics[width=0.95\textwidth]{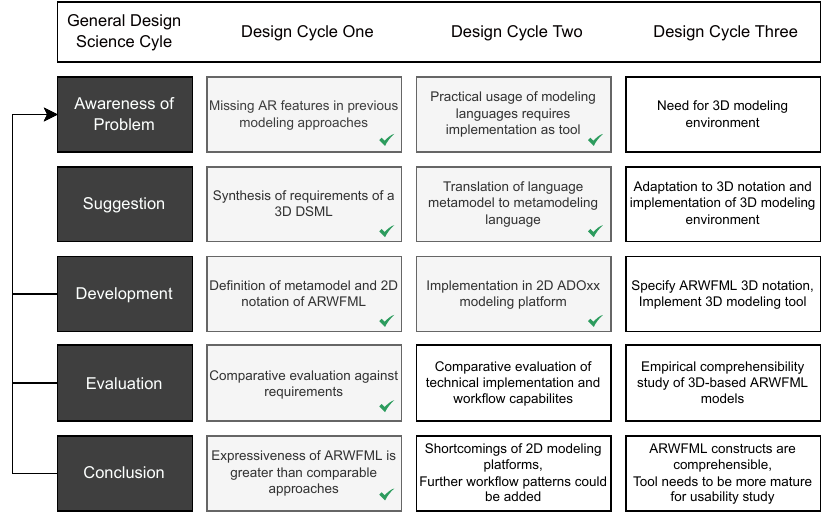}
    \caption{Design cycles for advancing the \textit{ARWFML} language~\cite{muff2023domain}, following \textit{Design Science Research Methods}~\cite{morana2019designing}. The light gray areas have already been addressed.}
    \label{fig:dsr_cycles}
\end{figure}

Therefore, in this paper, we continue to evaluate this previous implementation in comparison to similar approaches following design science research (DSR) methodologies~\cite{morana2019designing} -- see Fig.~\ref{fig:dsr_cycles}. Based on expert feedback received in this second design cycle, we conducted a comparison of the language's workflow capabilities with those of other languages. Subsequently, we will report on how we addressed the identified shortcomings in a third design cycle and finally arrived at a new implementation that is ready for assessing the comprehensibility of the language concepts in an empirical study. The main contributions of this paper are (i) detailed insights into the evaluation process of modeling languages for the domain of augmented reality by using \textit{ARWFML} as a concrete example, (ii) a comparison of technical features of AR modeling language implementations including their workflow capabilities, and (iii) the results from an empirical study for testing the comprehensibility of \textit{ARWFML} constructs. Thereby, we aim to inform how such evaluations can be accomplished for similar approaches.

The paper is structured as follows. Section \ref{sec:related_work} provides a brief revisit of \textit{ARWFML} and a comparative evaluation of its technical implementation to related approaches in Section~\ref{subsec:comparative_technical}. Section \ref{subsec:workflow_req} focuses on those approaches capable of creating AR workflow scenarios and compares them against the most relevant workflow patterns. This is followed by a description of the third design cycle in Section \ref{sec:comprehensibility}, which includes the design of a new three-dimensional notation for \textit{ARWFML} in Section~\ref{subsec:3D_notaton}, a newly developed, 3D-enhanced modeling environment presented in Section~\ref{subsec:m2ar}, and the results of an empirical comprehensibility study of the extended \textit{ARWFML} language in Section~\ref{subsec:empirical}. The paper ends with a conclusion and an outlook in Section~\ref{sec:conclusion}.

\section{Outline of ARWFML and Comparative Evaluation of Related Approaches in the Second Design Cycle}
\label{sec:related_work}

\begin{figure}[t!]
  \centering
  \includegraphics[width=\textwidth]{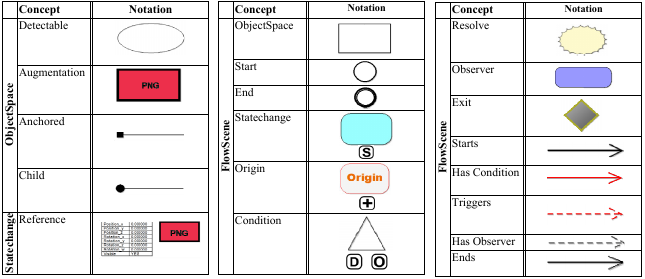}
  \caption{2D notation of of the \textit{ARWFML} language concepts, separated by the three different language types \textit{ObjectSpace}, \textit{Statechange}, and \textit{FlowScene}. }
  \label{fig:ARWFML_2D}
\end{figure}

The \textit{Augmented Reality Workflow Modeling Language} (ARWFML) is a domain-specific visual modeling language that allows the modeling of complex augmented reality workflows for different application scenarios~\cite{muff2023domain}. It includes three different types of visual models: \textit{ObjectSpace} models for defining the general AR environment, \textit{FlowScene} models for defining the AR workflow, and \textit{Statechange} models for specifying the different changes of virtual objects within this workflow. \textit{ARWFML}'s high level of abstraction frees users from the need to have technical expertise in AR application development, allowing them to focus on the content and functionality of AR scenarios. The created \textit{ARWFML} models are to be interpreted by an AR execution engine running AR experiences based on the loaded models. In a first design cycle, the \textit{ARWFML} metamodel has been specified and compared against requirements for model-based AR approaches. \textit{ARWFML} was shown to cover more requirements than any previous approach. Fig.~\ref{fig:ARWFML_2D} shows the visual elements of \textit{ARWFML}. In a second design cycle, the \textit{ARWFML} metamodel has been translated to the \textit{ADOxx} metamodeling language, and the language has been implemented on the 2D \textit{ADOxx} metamodeling platform by considering agile modeling method engineering principles~\cite{fill2013conceptualisation,buchmann2015agile}. 

\subsection{Comparative Evaluation of Technical Implementation}
\label{subsec:comparative_technical}

In the following, we continue with the evaluation of \textit{ARWFML} in the second design cycle. When evaluating modeling approaches for AR scenarios, it is not sufficient to consider only the language requirements for AR concepts as previously for \textit{ARWFML}~\cite{muff2023domain}. It is also necessary to consider how such modeling languages can be technically implemented, including the elicitation and execution of models to create AR scenarios and the necessary technical requirements. The reason for this is that the elicitation of models for 3D space, which is a prerequisite for AR scenarios, cannot be accomplished in 2D-based modeling environments. In addition, for languages that define workflows, it is important to verify the ability of a language to cover different workflow patterns~\cite{aalst2003workflow}. Ultimately, the question arises as to what methods can be used to evaluate the modeling approaches in more detail. Thus, for the evaluation part of the second design cycle, in the following, we conduct a comparative evaluation of the technical implementation, including the workflow capabilities, and evaluation types of the major related approaches compared to \textit{ARWFML} -- see Table~\ref{tab:language_comparison}. The comparison considers the dimensions \textit{Language Type}, \textit{Model Elicitation Environment}, \textit{Model Execution}, and \textit{Resulting Environment}. Furthermore, the approaches are compared in their capabilities to define workflows \textit{(Workflow)} and applied type of evaluation \textit{(Type of Evaluation)} -- see Table \ref{tab:language_comparison}.

The \textit{SSIML} and \textit{X3D} approach by Lenk et al. \cite{lenk2012model} investigated model-driven development and round-trip engineering for 3D web applications by converting text-based models to code for 3D experiences. Ruminski and Walczak~\cite{ruminski2014dynamic} and Lechner \cite{lechner2013arml} introduced the textual declarative languages \textit{CARL} and \textit{ARML 2.0} for modeling augmented reality experiences that can be interpreted by an execution engine or as browser standard and offer 3D environments for their specification. Grambow et al.~\cite{grambow2021context} presented the \textit{BPMN-CARX} approach, a visual extension of the BPMN language. Models can be interpreted in a run-time environment as AR experiences. Ruiz-Rube et al. \cite{ruiz_rube2020model} introduced \textit{ARE4DSL}, a text-based model-driven development approach for the generation of code that produces AR model editors. 
The resulting applications are geared towards modeling itself rather than developing AR applications in general.
Campos-Lopez et al. \cite{campos_l_opez2023model} proposed \textit{ALTER}, an automated textual approach to build augmented reality interfaces. Models are interpreted to run as a domain-specific language in an AR interface.
Lastly, Wild et al. \cite{wild2014towards,wild2020ieee} proposed an approach called \textit{ARLEM}, for the definition of AR learning activities. A reference implementation allows for immersive model creation in three-dimensional space and the runtime interpretation in an AR application.

\begin{table}[ht]
\tiny
\centering
\caption{Comparison table for related AR modeling approaches. (D): Domain-specific languages. (P): Language profiles. 
(Y): Dimension covered by approach. (N): Dimension not covered by approach.}
\begin{tblr}{
  colsep = 3pt,
  column{even} = {c},
  column{3} = {c},
  column{5} = {c},
  column{7} = {c},
  column{9} = {c},
  hline{1-2,23} = {-}{},
  hline{4,9,12,15,17} = {-}{dotted},
}
    & {\textbf{SSIML/}  \\\textbf{X3D}  \\\cite{lenk2012model}} 
    & {\textbf{CARL}    \\{\mbox{}}                \\\cite{ruminski2014dynamic}} 
    & {\textbf{BPMN-}   \\\textbf{CARX} \\\cite{grambow2021context}} 
    & {\textbf{ARML}    \\{\mbox{}}                \\\cite{lechner2013arml}} 
    & {\textbf{ARE4DSL} \\{\mbox{}}                \\\cite{ruiz_rube2020model}} 
    & {\textbf{ALTER}   \\{\mbox{}}                \\\cite{campos_l_opez2023model}} 
    & {\textbf{ARLEM}   \\{\mbox{}}                \\\cite{wild2020ieee}} 
    & {\textbf{ARWFML}  \\{\mbox{}}                 \\\cite{muff2023domain}} \\
\textbf{Language Type}                                  &                                 &               &                                 &               &                  &                &                &                 \\
DSML/Profile                                                & D                               & D             & P                               & D             & D                & D              & D              & D               \\
{\textbf{Model Elicitation}\\\textbf{Environment}} &                                 &               &                                 &               &                  &                &                &                 \\
Textual                                            & Y                               & Y             & Y                               & Y             & Y                & Y              & Y              & Y               \\
Visual 2D                                          & N                               & N             & Y                               & N             & N                & N              & N              & Y               \\
Visual 3D                                          & N                               & N             & N                               & N             & N                & N              & N              & N               \\
Immersive                                          & N                               & N             & N                               & N             & N                & N              & Y              & N               \\
\textbf{Model Execution}                           &                                 &               &                                 &               &                  &                &                &                 \\
Code Generation                                    & Y                               & N             & N                               & N             & Y                & N              & N              & N               \\
Model Interpretation                               & N                               & Y             & Y                               & Y             & N                & Y              & Y              & Y               \\
{\textbf{Resulting}\\\textbf{Environment}}         &                                 &               &                                 &               &                  &                &                &                 \\
3D Experience                                      & Y                               & Y             & Y                               & Y             & N                & N              & Y              & Y               \\
Modeling Environment                               & N                               & N             & N                               & N             & Y                & Y              & Y              & N               \\
\textbf{Workflow}                                  &                                 &               &                                 &               &                  &                &                &                 \\
Workflow Execution                                 & N                               & N             & Y                               & N             & N                & N              & Y              & Y               \\
\textbf{Type of Evaluation}                        &                                 &               &                                 &               &                  &                &                &                 \\
Use Case Demonstration                                      & Y                               & Y             & Y                               & Y             & Y                & Y              & Y              & Y               \\
Implementation                                     & Y                               & Y             & Y                               & N             & Y                & Y              & Y              & Y               \\
{Empirical Comprehen-\\sibility Study}             & N                               & N             & N                               & N             & N                & N              & N              & N               \\
{Empirical \\Tool Usability Study}                 & N                               & N             & N                               & N             & Y                & Y              & N              & N               \\
Empirical Simulation                               & N                               & N             & Y                               & N             & N                & N              & N              & N               
\end{tblr}
\label{tab:language_comparison}
\end{table}
\normalsize

Only the three approaches \textit{BPMN-CARX} \cite{grambow2021context}, \textit{ARLEM} \cite{wild2020ieee} and \textit{ARWFML} \cite{muff2023domain} offer workflow capabilities for specifying complex branching in AR experiences. Regarding the type of evaluation considered by the different approaches, all of them demonstrated use cases. All approaches, except \textit{ARML 2.0} \cite{lechner2013arml}, have been evaluated with a first implementation as a proof of concept. No approach evaluated the comprehensibility of the proposed language, and only \textit{ARE4DSL} \cite{ruiz_rube2020model} and \textit{ALTER} \cite{campos_l_opez2023model} have been evaluated regarding the usability of the proposed prototype. In addition, \textit{BPMN-CARX} \cite{grambow2021context} has been evaluated by conducting an empirical simulation.
\textit{ARWFML} has also been implemented and use cases have been shown on the basis of the two-dimensional \textit{ADOxx} metamodeling platform \cite{fill2013conceptualisation}. However, the second design cycle showed that a further evaluation of the workflow requirements is necessary on the level of the \textit{ARWFML} metamodel. Furthermore, the language should be empirically evaluated by users to assess its practical feasibility.
Therefore, in the next section, we advance to an evaluation of \textit{ARWFML} against common workflow patterns, which will be followed by the description of the third design cycle to establish the foundations for a first empirical evaluation in Section \ref{sec:comprehensibility}. 

\subsection{Evaluation of Workflow Capabilities} \label{subsec:workflow_req}

Modeling languages for depicting and executing workflows have a long tradition. They permit to structure sequences of activities to complete tasks and may support different types of patterns, which characterize the expressiveness of a language. The most significant patterns were summarized by van der Aalst et al.~\cite{van_der_aalst2002w,aalst2003workflow}. Typical examples for workflow modeling languages include Petri nets \cite{petri2008petri} or BPMN\footnote{\url{https://www.omg.org/spec/BPMN/2.0/} last visited on: 25.04.2024}. 

In the following, we compare \textit{ARWFML} against two other AR modeling languages that enable the specification of workflows based on 20 core workflow patterns~\cite{aalst2003workflow}. Various additional patterns exist. However, van der Aalst et al. mention that these 20 are the most significant patterns~\cite{van_der_aalst2002w}. Furthermore, we present a use case for an \textit{ARWFML} \textit{FlowScene} model that illustrates all the workflow patterns supported by \textit{ARWFML}. Based on this analysis, we derive further language requirements that will be considered in subsequent design iterations. Table~\ref{tab:workflow_patterns} shows a summary of the modeling languages \textit{BPMN-CARX}, \textit{ARLEM}, and \textit{ARWFML} regarding 20 selected workflow patterns. 
\begin{table}[t!]
\tiny
\centering
\caption{Comparison of support of workflow patterns according to \cite{aalst2003workflow} by three AR modeling languages. (Y): Full support. (N): No support. ($\sim$): Partial support.}
\begin{tblr}{
  colsep = 3pt,
  column{even} = {c},
  column{3} = {c},
  hline{1,29} = {-}{0.08em},
  hline{2,28} = {-}{0.05em},
  hline{8,13,16,21,25} = {-}{dotted},
}
\textbf{Workflow Pattern}                                                                     & {\textbf{BPMN-CARX}\\\cite{grambow2021context}}         & {\textbf{ARLEM}\\\cite{wild2020ieee}} & {\textbf{ARWFML}\\\cite{muff2023domain}} \\
\textbf{Basic Control Flow Patterns}                                           & ~                 & ~     & ~      \\
Pattern 1
  (Sequence)                                                & Y                 & Y     & Y      \\
Pattern 2
  (Parallel Split)                                          & Y                 & N     & Y      \\
Pattern 3
  (Synchronization)                                         & Y                 & N     & N      \\
Pattern 4
  (Exclusive Choice)                                        & Y                 & Y     & N      \\
Pattern 5
  (Simple Merge)                                            & Y                 & Y     & Y      \\
{\textbf{Advanced Branching and Synchronization Patterns}}             & ~                 & ~     & ~      \\
Pattern 6
  (Multi - choice)                                          & Y                 & N     & Y      \\
Pattern 7
  (Synchronizing Merge)                                     & $\sim$ & N     & N      \\
Pattern 8
  (Multi - merge)                                           & Y                 & N     & Y      \\
Pattern 9
  (Discriminator)                                           & $\sim$ & N     & N      \\
\textbf{Structural Patterns}                                                 & ~                 & ~     & ~      \\
Pattern 10
  (Arbitrary Cycles)                                       & Y                 & Y     & Y      \\
Pattern 11
  (Implicit Termination)                                   & Y                 & N     & Y      \\
\textbf{Patterns involving Multiple Instances}                               & ~                 & ~     & ~      \\
Pattern 12
  (Multiple Instances Without Synchronization)             & Y                 & N     & N      \\
Pattern 13
  (Multiple Instances With a Priori Design Time Knowledge) & Y                 & N     & N      \\
Pattern 14
  (Multiple Instances With a Priori Runtime Knowledge)     & Y                 & N     & N      \\
Pattern 15
  (Multiple Instances Without a Priori Runtime Knowledge)  & N                 & N     & N      \\
\textbf{State-based Patterns}                                                & ~                 & ~     & ~      \\
Pattern 16
  (Deferred Choice)                                        & Y                 & N     & N      \\
Pattern 17
  (Interleaved Parallel Routing)                           & $\sim$ & N     & N      \\
Pattern 18
  (Milestone)                                              & N                 & N     & N      \\
\textbf{Cancellation Patterns}                                               & ~                 & ~     & ~      \\
Pattern 19
  (Cancel Activity)                                        & Y                 & N     & N      \\
Pattern 20
  (Cancel Case)                                            & Y                 & N     & Y \\
  \textbf{$\sum$ (fully supported)} & \textbf{15} & \textbf{4} & \textbf{8}
\end{tblr}
\label{tab:workflow_patterns}
\end{table}
\normalsize
The analysis reveals significant differences in the workflow capabilities of each modeling language. \textit{BPMN-CARX} shows the most comprehensive support, successfully implementing 15 of the 20 analyzed patterns. This includes comprehensive support for all basic control flow patterns and the majority of advanced branching, synchronization, and multiple instances patterns, indicating a flexible approach to complex workflow scenarios. Since \textit{BPMN-CARX} is an extension of the BPMN modeling language, it comes with the extensive capabilities of the base language and allows for very complex workflows. Conversely, the language is constrained by its focus on process flows in general, with only basic annotations of tasks with AR functionalities. It is not tailored for the definition of spatial models, whereas the other two approaches are explicitly developed for defining workflows for AR.

In contrast, \textit{ARLEM} supports only four patterns. Three basic control flow patterns, and arbitrary cycles. This suggests a focus on simpler, more linear workflows without the need for complex synchronization. \textit{ARLEM}, as well as \textit{ARWFML} do not support patterns involving multiple instances and state-based decisions. Multiple instance patterns describe situations where there are multiple threads of execution active in a process model, each relating to the same activity \cite{aalst2003workflow}. Given that there can be only one instance of an AR experience at a given time, these patterns are not considered in \textit{ARLEM}, nor \textit{ARWFML}.
The limited support of workflow patterns can restrict the use of \textit{ARLEM} in more complex or variable workflow scenarios. It is important to note that \textit{ARLEM} is designed to define learning experiences and does not claim to allow the definition of highly complex scenarios.

\begin{figure}[t!]
    \centering
    \includegraphics[width=\linewidth]{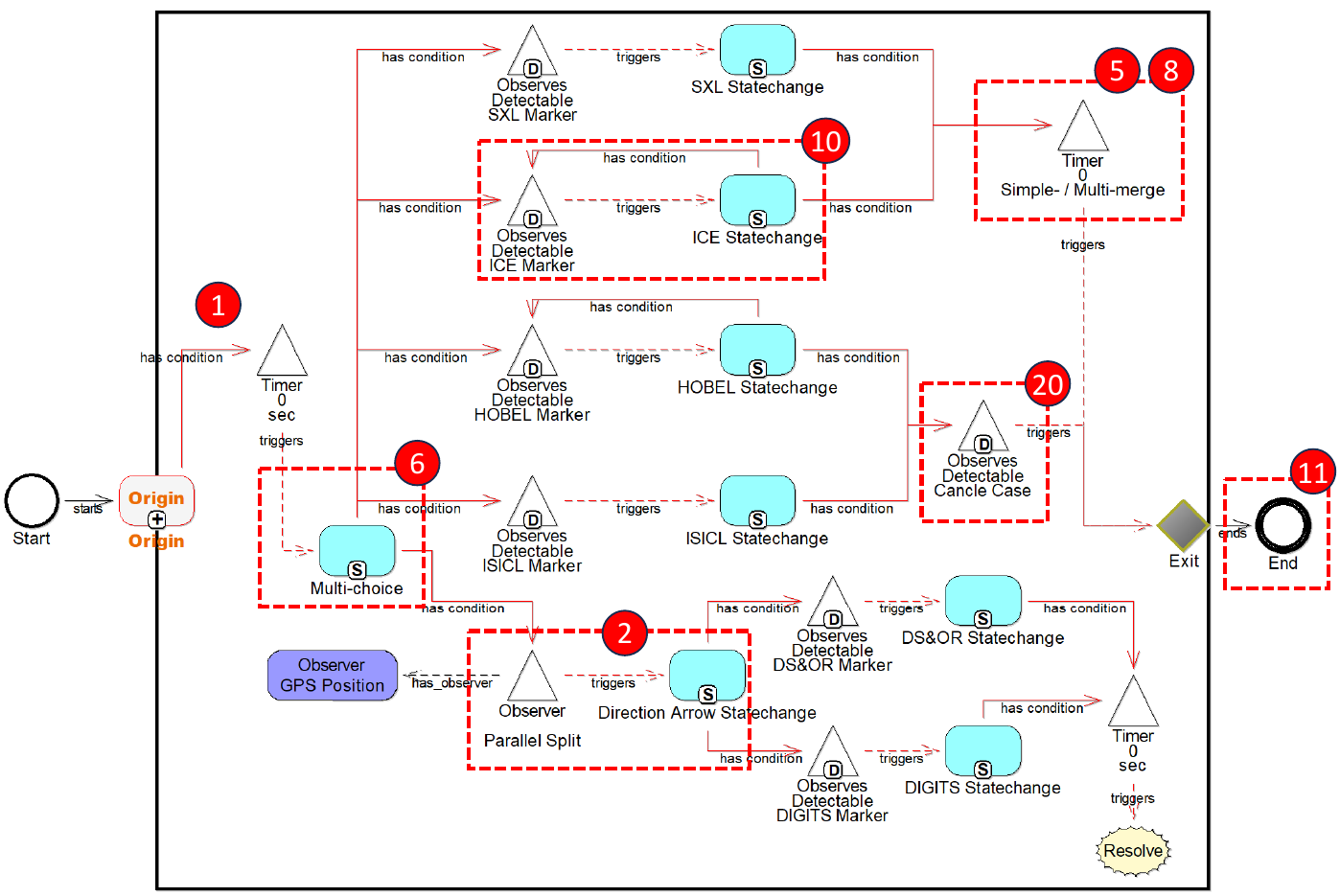}
    \caption{\textit{ARWFML FlowScene} model illustrating the eight supported workflow patterns by \textit{ARWFML}. Each pattern in the model is marked with a red circle indicating the number of the workflow pattern according to Table \ref{tab:workflow_patterns}. (1) \textit{Sequence}; (2) \textit{Parallel Split}; (5) \textit{Simple Merge}; (6) \textit{Multi-choice}; (8) \textit{Multi-merge}; (10) \textit{Arbitrary Cycles}; (11) \textit{Implicit Termination}; (20) \textit{Cancel Case}.}
    \label{fig:FlowScene_example_adoxx}
\end{figure}

ARWFML supports eight workflow patterns, including the most basic control flow patterns, which are essential for straightforward sequential and parallel processes. This includes the \textit{Sequence}, \textit{Parallel Split}, and \textit{Simple Merge} patterns. However, it shows limitations in handling more complex synchronization and exclusive choice patterns, which are critical for dynamic decision-making scenarios in workflows. In advanced branching and synchronization, \textit{ARWFML} supports patterns like \textit{Multi-choice} and \textit{Multi-merge}. It lacks support for the \textit{Synchronizing Merge} and \textit{Discriminator} patterns, indicating some limitations in splitting and merging multiple concurrent paths within a workflow. The language's support extends to structural patterns, particularly handling arbitrary cycles and implicit terminations effectively. As stated previously, it is not feasible to execute multiple instances of an AR workflow on a single device concurrently, given that there is only one AR session running at a time. Thus, patterns involving multiple instances are not considered. 
Furthermore, state-based patterns such as \textit{Deferred Choice}, \textit{Interleaved Parallel Routing}, or \textit{Milestone} are not yet supported by \textit{ARWFML}. Cancellation is possible through the \textit{Cancel Case} pattern. Given that \textit{ARWFML} does not have the concept of activities, the \textit{Cancel Activity} pattern is not applicable.

To demonstrate the workflow patterns supported by \textit{ARWFML}, Fig. \ref{fig:FlowScene_example_adoxx} shows an example of a \textit{FlowScene} model. All supported workflow patterns are indicated with the corresponding pattern number in the model. The use case presented in this example depicts a scenario in which an AR user has the potential to visit various research groups on a university campus. At the physical location of each research group on the campus, a marker is placed. As soon as the user comes within proximity to the marker, visual information about the research group is displayed in AR. There are various conditions and paths that can be activated, such as the initiation of a flow, only if the user is within the specified range of \textit{GPS} coordinates. Some areas may be revisited multiple times (\textit{Arbitrary Cycle}), while others may only be visited once. 

In summary, \textit{BPMN-CARX} stands out as the most versatile and capable language, especially for environments where complex, multi-faceted workflows are common, but it lacks simplicity since it is based on the BPMN modeling language. \textit{ARLEM} might be more suited for simpler and less dynamic environments. \textit{ARWFML} provides a robust foundation for fundamental and some advanced workflow scenarios. However, further development may be necessary to fully support the more complex and dynamic patterns required in highly variable and decision-intensive workflows. 
For example, by modifying the \textit{ARWFML} metamodel to allow exclusive choices in the condition concept and introducing the possibility of defining synchronization conditions, patterns 3, 4, 7, and 9 could be supported. Furthermore, by extending the observer concept with incoming messages, pattern 16 could be supported. This would result in the number of supported patterns increasing to 13, thereby approaching the number of patterns that can be supported by the BPMN modeling language. Thus, we will tackle these changes in a fourth design cycle. Before doing so, in the next section, we will advance with a first empirical evaluation of the comprehensibility of the language in its current form by users and show the corresponding challenges. 

\section{Third Design Cycle: Towards Empirical Evaluations}
\label{sec:comprehensibility}

As discussed in Section \ref{sec:related_work}, different paths can be taken to evaluate AR modeling languages. The most common approaches include use case demonstrations, e.g., \cite{lenk2012model,ruminski2014dynamic,grambow2021context,lechner2013arml,ruiz_rube2020model,campos_l_opez2023model,wild2020ieee} and proof of concept implementations, e.g., \cite{lenk2012model,ruminski2014dynamic,grambow2021context,ruiz_rube2020model,campos_l_opez2023model,wild2020ieee}. When considering the involvement of users, recent approaches conducted usability \cite{ruiz_rube2020model,campos_l_opez2023model}, or simulation studies \cite{grambow2021context}.
Simulation studies and usability studies are strongly dependent on software tools. As a result, they do not evaluate the modeling language itself, but rather a combination of the modeling language and the implementation in the tool. Thus, the evaluation of the comprehensibility of the modeling language should be first conducted independently of the software tool, e.g., as proposed by Bork et al. \cite{bork2019intuitive}.

However, since \textit{ARWFML} is a domain-specific modeling language that allows for the modeling of 3D \textit{ObjectSpaces} and \textit{Augmentations} therein, the previously developed 2D notation was found to be inadequate~\cite{muff2023domain}. 
As a consequence, it was necessary to design a new three-dimensional notation of the \textit{ARWFML} language constructs.
With the help of this 3D notation, models that are more adequate for an empirical evaluation of the comprehensibility can be provided. The question of how these models are created arises. In principle, mock-up tools (e.g., Figma\footnote{\url{https://www.figma.com/community}}), image editing software (e.g., Adobe Photoshop\footnote{\url{https://www.adobe.com/products/photoshop.html}}), or 3D scene editing tools (e.g., Blender\footnote{\url{https://www.blender.org/}}) could be used. However, the definition of valid conceptual models with these purely graphical tools poses a challenge, as the creation may easily lead to formal inaccuracies and inconsistencies, and thus invalid models. 
Consequently, for the purpose of conducting a user-based comprehensibility study, it was necessary to develop a novel 3D-enabled modeling environment that facilitates the generation of formally accurate models in 3D notation.


\begin{figure}[t!]
  \centering
  \includegraphics[width=\textwidth]{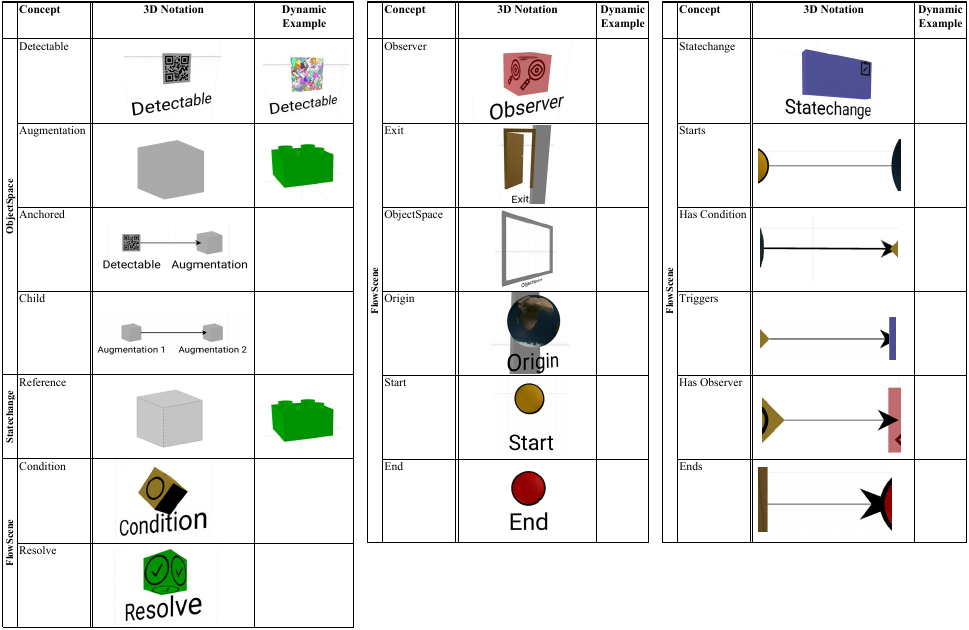}
  \caption{3D notation of of the \textit{ARWFML} language concepts grouped by the \textit{SceneTypes} \textit{ObjectSpace}, \textit{Statechange}, and \textit{FlowScene}. For each \textit{SceneType}, the visual notation is shown in the \textit{3D Notation} column. If there is a dynamic visualization depending on attribute values, an example visualization appears in the \textit{Dynamic Example} column.}
  \label{fig:ARWFML_3D}
\end{figure}

\subsection{Preliminaries: Adaptation of ARWFML to 3D} \label{subsec:3D_notaton}
In the third iteration of the DSR research project, the 2D notation of \textit{ARWFML} (see Fig. \ref{fig:ARWFML_2D}) has been adapted to a 3D notation (see Fig. \ref{fig:ARWFML_3D}). It is important to note that \textit{ARWFML} contains three-dimensional dynamic visualizations for the classes Detectable, \textit{Augmentation}, and Reference. Fig. \ref{fig:ARWFML_3D} shows examples of dynamic visualizations of these classes in column ``Dynamic Example''. For the \textit{Detectables}, an illustrative example of a special color marker that can be identified by computer vision algorithms is provided. As an illustration of an \textit{Augmentation}, a 3D object in the form of a green color brick that can be 3D visualized in an AR application is presented. As the Reference class is linked to \textit{Augmentations}, the dynamic visualization of References displays the same 3D model as the referenced \textit{Augmentation}, in this example also the green color brick.

\subsection{The 3D Modeling Environment M2AR}
\label{subsec:m2ar}

In order to facilitate the creation of 3D-enhanced \textit{ARWFML} models, a new modeling environment, designated as \textit{M2AR}, has been developed. \textit{M2AR} is a web-based modeling environment that enables the creation of traditional 2D models as well as models in three-dimensional space~\cite{muff2024m2ar}. The platform is based on the 3D JavaScript library \textit{THREE.js}\footnote{\url{https://github.com/mrdoob/three.js/}} and the mixed reality immersive web standard WebXR~\cite{jones2023webxr}. 

\begin{figure}[t!]
    \centering
    \includegraphics[width=.85\linewidth]{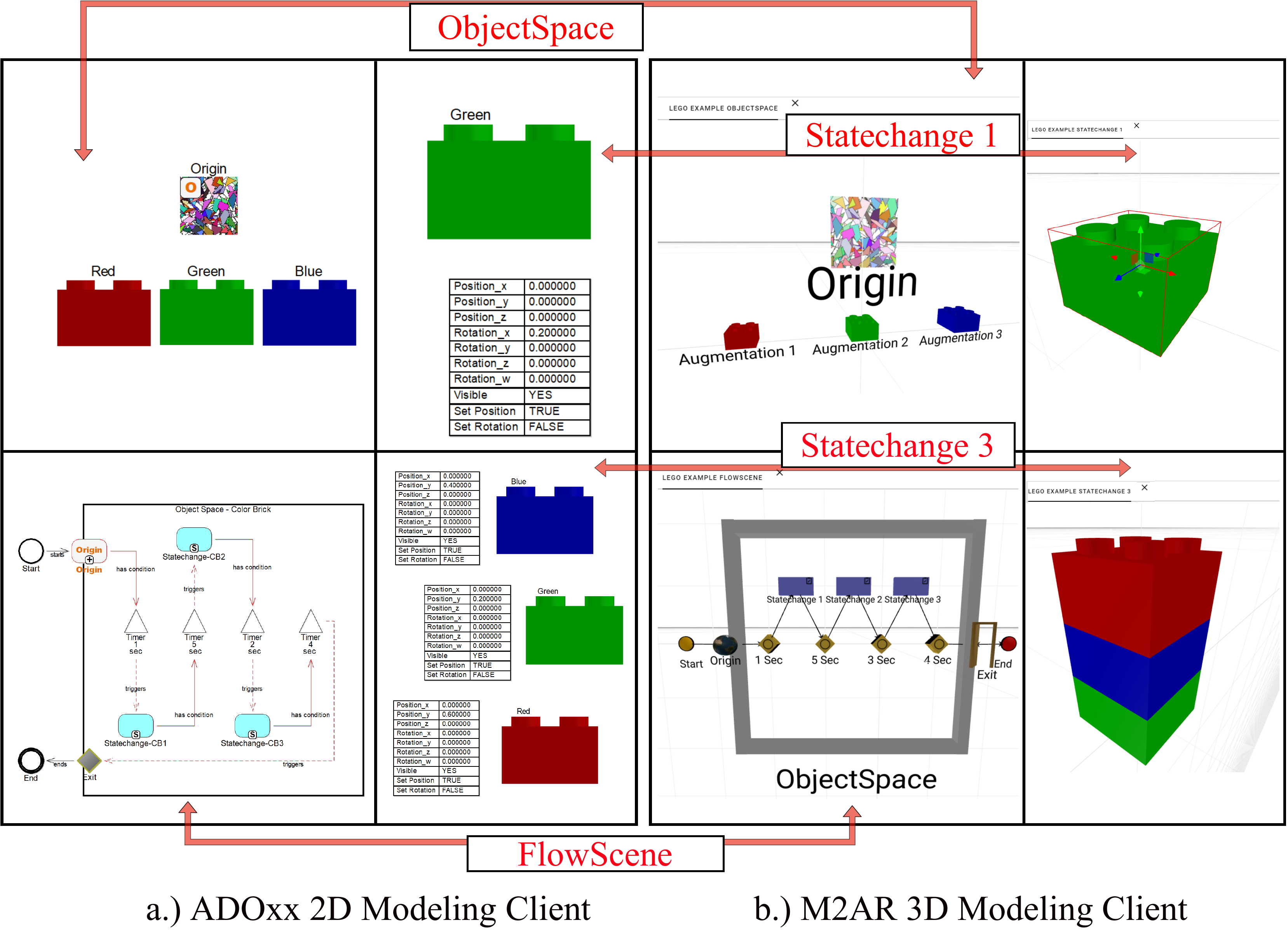}
    \caption{Visual comparison of \textit{ObjectSpace}, \textit{Statechange}, and \textit{FlowScene} models in  (a.) 2D notation of \textit{ARWFML} in the \textit{ADOxx} modeling tool from the first design cycle and (b.) the new 3D notation in the new \textit{M2AR} modeling client.}
    \label{fig:2D_3D_comparison}
\end{figure}

On this new platform, the novel 3D notation for \textit{ARWFML} was implemented in a way that facilitates spatial alignment of disparate three-dimensional objects. This is particularly advantageous when modeling \textit{Statechanges}, as they define the manner in which the various \textit{Augmentations} are represented in the real world. This addresses the necessity for an environment that permits three-dimensional spatial modeling, derived at the conclusion of the initial design cycle described in \cite{muff2023domain}. Fig. \ref{fig:2D_3D_comparison} illustrates a comparison of \textit{ARWFML} model excerpts of the 2D notation modeled in the \textit{ADOxx} implementation from the second design cycle (cf. \cite{muff2023domain}) and the 3D notation in the new \textit{M2AR} modeling environment. 

\textit{M2AR} is a prototypical modeling environment that has not yet reached a level of maturity sufficient to empirically evaluate its usability due to the high expectations users have for such tools today. However, it allows expert modelers to create valid \textit{ARWFML} models that can then be interpreted by an \textit{AR Engine} that takes models from the 3D modeling client as input to execute modeled AR applications. Consequently, models created with \textit{M2AR} are well-suited for an empirical evaluation of the comprehensibility of the modeling language. 

\subsection{Empirical User Study for ARWFML Model Comprehensibility}
\label{subsec:empirical}

This section presents the design and findings of an empirical user study conducted to assess the comprehensibility of \textit{ARWFML} in 3D notation. We describe the study design, followed by data on the experimental subjects, the evaluation metrics used, the study results, and a discussion of potential threats to the study's internal and external validity.


\textbf{Study Design}: To evaluate the comprehensibility of \textit{ARWFML}, our study employs a between-subject design \cite{charness2012experimental}. Thereby, two user groups were independently tested. This design was chosen to identify differences between user groups with different backgrounds. The main threats to the validity of this setup will be addressed later. The experiments were conducted 
at the end of 2023.

Each study consisted of two parts: (1) an introduction to the \textit{Augmented Reality Workflow Modeling Language}, which covered the basic concepts of AR, the specific \textit{ARWFML} language concepts, and an introduction to the new \textit{M2AR} environment featuring the 3D notation. The introduction included a 45-minute presentation with examples of the different language concepts of \textit{ARWFML}. Following the presentation, participants were asked to complete a questionnaire regarding their demographics, understanding of \textit{ARWFML} models, and general comprehension of the different \textit{ARWFML} language concepts. To assess model understanding, participants were shown five sections of screenshots from \textit{ARWFML} models created with \textit{M2AR}, each with three different possibilities for the resulting workflow in the resulting AR application. The participants were then asked to choose the correct result for each question.

\begin{table}[t]
\scriptsize
\centering
\caption{Categories of survey question (Q1 - Q5) for evaluating \textit{ARWFML} model understanding.}
\label{tab:ARWFML_questions_q}
\begin{tblr}{
  colspec={X[0.04] X[0.15] X[0.70]}, 
  hline{1-2,7} = {-}{},
  vline{2-3} = {-}{dotted},
}
Q  & Use Case & Purpose                     \\
Q1 & Color Brick & Prediction of outcome showing different Statechanges. \\
Q2 & Color Brick & Prediction of outcome when differing Conditions. \\
Q3 & Solar System & Prediction of outcome showing different Statechanges.\\
Q4 & Solar System & Prediction of outcome with different Statechanges and Conditions.\\
Q5 & Industrial Machine & Prediction of outcome showing different Statechanges.\\
\end{tblr}
\end{table}
\normalsize

The purpose of the questions was to assess participants' comprehension of the visual models and determine the predictability of the model outcomes. Table~\ref{tab:ARWFML_questions_q} shows the categories of questions related to model understanding, including the use case represented and the purpose of each category\footnote{The survey is available on: \url{https://doi.org/10.5281/zenodo.13219036}}. A total of three use cases were included. The first use case guides potential users through the assembly of different colored bricks. The second use case, the solar system use case, provides a learning opportunity by displaying the different planets and moons in our solar system. The third use case, the industrial machine use case, guides users through a working process, indicating which buttons to press. 


Additionally, participants were presented with statements about the \textit{ARWFML} modeling language and asked to identify all correct statements in five separate questions (see general questions GQ1-GQ5 in Table \ref{tab:ARWFML_questions}). The purpose of the general questions was to gather information about the specific concepts of the modeling method. GQ1 had the goal of determining if participants could recall the different \textit{ARWFML} \textit{SceneTypes} presented in a selection of unrelated other \textit{SceneTypes}. GQ2 aimed to determine whether participants were able to recall the purpose of the three different \textit{SceneTypes} by selecting true statements about each one. Questions GQ3 to GQ5 were presented separately for each \textit{SceneType} of \textit{ARWFML}. The participants were presented with different choices of different concepts. They had to select only the concepts that are part of the \textit{SceneType} to which the question refers. The purpose of these questions was to evaluate the participant's understanding of the \textit{SceneTypes} of the \textit{ARWFML}.

\begin{table}[t!]
\scriptsize
\centering
\caption{General questions (GQ1 - GQ5) for evaluating \textit{ARWFML} method understanding.}
\label{tab:ARWFML_questions}
\begin{tblr}{
  colspec={X[0.07] X[0.74] X[0.19]}, 
  hline{1-2,7} = {-}{},
  vline{2-3} = {-}{dotted},
}
GQ  & Question                                                                                                                   & Purpose                     \\
GQ1 & Which are the 3 different types of models that can be created with ARWFML?                                                  & SceneTypes                \\
GQ2 & {ARWFML defines the three types ObjectSpace, Statechange, and FlowScene. Choose the three right statements.}                       & {SceneType\\Purpose}        \\
GQ3 & {Which concepts are available in an ObjectSpace model? Choose the two available modeling concepts of an ObjectSpace model.}   & {ObjectSpace\\Understanding} \\
GQ4 & {Which concepts are available in a Statechange model? Choose the only available modeling concepts of a Statechange model.} & {Statechange\\Understanding} \\
GQ5 & {Which concepts are available in a FlowScene model? Choose the seven introduced modeling concepts of a FlowScene model.}       & {FlowScene\\Understanding}   
\end{tblr}
\end{table}
\normalsize

The sections that follow will cover the experimental subjects, evaluation metrics, and questionnaire results in quantitative terms.

\textbf{Experimental Subjects}: To ensure that the subjects had at least a basic understanding of modeling, we recruited them from university courses in ``Introduction to Business Informatics'' and ``Databases'' at the University of Fribourg, where various modeling methods are taught. Our sample consisted of 35 students, aged between 18 and 44 years, the majority of whom were under 24 years of age. 
The study involved participants with varying levels of education, including high school graduates (25), individuals with Bachelor's degrees (7), Master's degrees (2), and Doctoral degrees (1).

To evaluate the participants' familiarity with \textit{conceptual modeling} and \textit{augmented reality}, we gathered data using five-point Likert scales \cite{likert1932technique}. 
Responses are categorized as ``Very Familiar'', ``Familiar'', ``Somewhat familiar'', ``Somewhat unfamiliar'', and ``Very unfamiliar'', with corresponding participant counts for each category. The data indicate a range of exposure and understanding among the participants in these two domains. Of the 35 respondents, 18 rated their familiarity with \textit{conceptual modeling} as ``Somewhat familiar'' or higher, and four considered themselves ``Familiar''. In the context of \textit{augmented reality}, 25 participants stated that they are ``Somewhat unfamiliar'' or ``Very unfamiliar''.


\textbf{Evaluation Metrics}: Measuring the comprehensibility of a modeling method is difficult, given that modeling is typically performed using a modeling tool, making it challenging to separate the method from the tool for evaluation. Moody's \textit{Methods Evaluation Model} \cite{moody2003method} defines four measures to considered when evaluating information systems design methods in general:
\textit{Performance}, \textit{Perceptions}, \textit{Intentions}, and \textit{Behavior}. Since we wanted to objectively evaluate only the modeling language that is not yet used in practice, we only evaluated \textit{Performance}, which is composed by the measures \textit{Actual Efficiency} and \textit{Actual Effectiveness} \cite{kekes1994pragmatic}. The efficiency of a method is determined by the amount of effort required to apply it. The effectiveness of a method is determined by how well it reaches its objectives. To evaluate the performance of an unproductive method, we have simplified the assessment to determine whether participants can distinguish between different model scenarios and understand the different language concepts. This is achieved by presenting them with various models for validation and asking general questions about the methodology (see Table \ref{tab:ARWFML_questions}).



\textbf{Results}: This section presents the results of the empirical comprehensibility study. The results for Q1-Q5 are discussed first, followed by the results of GQ1-GQ5. Additionally, a correlation analysis for Q1-Q5 and GQ1-GQ5 in regard to familiarity with conceptual modeling and augmented reality is presented.

\textbf{Q1-Q5}: Table~\ref{tab:accuracy_Q1-Q5} shows the percentage of correct answers for each of the five questions on \textit{ARWFML} model understanding (Q1 to Q5) in the two studies. The data is presented to show the percentage of participants choosing the right answer out of three given possibilities. 
A ``Total'' row aggregates the performance of both studies, providing an overall success rate for each question.

The results show a high overall percentage of correct responses for Q1 (0.94), Q4 (0.94), and Q5 (0.97) on average, indicating a robust understanding of the models. Q2 had moderate accuracy (0.91) with a notable increase in study group two. Q3 had the lowest average accuracy (0.83), which is still moderate. These findings indicate that the participants understand the models presented and the intended AR application scenario derived from these models.

\textbf{GQ1-GQ5}: Table \ref{tab:avg_GQ1-5} shows the average accuracy of responses across the two studies for the five general question on \textit{ARWFML} method understanding (GQ1- GQ5) -- refer to Table \ref{tab:ARWFML_questions}. The questions were formatted as multiple-choice, with different amounts of possible answers and correct responses. The bottom row displays the total average value for correct answers over all study participants. These five questions provide insights into the relative difficulty or clarity of the different concept of the modeling language (see \textit{Purpose} column in Table \ref{tab:ARWFML_questions}).


\begin{table}[t]
\scriptsize
\begin{minipage}{0.45\textwidth}
\centering
\caption{Percentage of correct responses across studies for questions on model understanding (Q1 to Q5).}
\label{tab:accuracy_Q1-Q5}
\begin{tblr}{
  row{even} = {c},
  row{3} = {c},
  row{5} = {c},
  hline{1,7} = {-}{0.08em},
  hline{2} = {-}{0.05em},
}
   & Study 1 & Study 2 & Total \\
Q1 & 0.94    & 0.95    & 0.94  \\
Q2 & 0.88    & 0.95    & 0.91  \\
Q3 & 0.81    & 0.84    & 0.83  \\
Q4 & 0.94    & 0.95    & 0.94  \\
Q5 & 1.00    & 0.95    & 0.97  
\end{tblr}
\end{minipage}\hfill
\begin{minipage}{0.45\textwidth}
\centering
\caption{Average correct answer rate for questions on \textit{ARWFML} method understanding (GQ1 to GQ5).}
\label{tab:avg_GQ1-5}
\begin{tblr}{
row{even} = {c},
  row{3} = {c},
  row{5} = {c},
  hline{1,7} = {-}{0.08em},
  hline{2} = {-}{0.05em},
}
    & Study 1 & Study 2 & Total \\
GQ1 & 0.91    & 0.88    & 0.89  \\
GQ2 & 0.78    & 0.87    & 0.83  \\
GQ3 & 0.90    & 0.92    & 0.91  \\
GQ4 & 0.92    & 0.86    & 0.89  \\
GQ5 & 0.84    & 0.79    & 0.81  
\end{tblr}
\end{minipage}
\end{table}

The results of both studies indicated that question GQ3 (\textit{ObjectSpace} Understanding) consistently achieved the highest and second-highest average scores for correct responses (scores of 0.90 and 0.92), suggesting that it was possibly the simplest or most comprehensible language concept.
The lowest average correct answer rate in the first study was observed for GQ2 (SceneType Purpose) at 0.78, while in study two it was observed for GQ5 (\textit{FlowScene} Understanding) at 0.79. This suggests that these concepts were the most difficult or least clear.
In the second group, the mean correct answer rates for questions GQ2 and GQ3 were generally higher than in the first group, but lower for questions GQ1, GQ4, and GQ5.
The total average correct answer rate across both groups for each question shows that, overall, GQ3 (Object Space model) had the highest average correct answer rate (0.91), while GQ2 (SceneType Purpose) had the lowest (0.83). This indicates that the \textit{ObjectSpace} concept was the best understood or easiest language concept (GQ2), while it is more challenging for participants to distinguish the general purpose of the different \textit{SceneTypes} (GQ3).

\textbf{Correlation}: To investigate the performance disparities related to varying degrees of familiarity with conceptual modeling or AR, we calculated the Pearson correlation coefficients between the measures of familiarity and the rates of correct answers for various \textit{ARWFML} questions, based on a five-point Likert scale. Age and education level were not included in the correlation analysis because most of the participants fell into the same categories. 
The correlation coefficients were all in the range between $-0.12$ and $0.39$ (see Table \ref{tab:correlation}). 
A t-test was performed to assess the statistical significance of the observed results. The analysis revealed that none of the values tested exhibited a significant correlation at the significance level $\alpha = 0.01$. However, GQ1 was the only question that demonstrated a significant correlation between familiarity with conceptual modeling and the correct answer rate at the significance level $\alpha = 0.05$.  It is important to highlight that the study involved only 35 participants, which could affect the robustness and generalizability of the correlation analysis results.

\begin{table}[t!]
\scriptsize
\centering
\caption{
Correlation coefficients between correct answer rates for \textit{ARWFML} questions and familiarity measures for conceptual modeling or augmented reality. (*) Significant correlation at $\alpha = 0.05$.}
\begin{tabular}{l l l l l l l}  
\toprule
 { } & Q1-Q5 & GQ1 & GQ2 & GQ3 & GQ4 & GQ5 \\
\midrule
Conceptual Modeling & { 0.02} & { 0.39*} & { 0.32} & {-0.12} & { 0.07} & { 0.33} \\
Augmented Reality & {-0.11} & { 0.31} & { 0.08} & { 0.11} & { 0.22} & { 0.19} \\
\bottomrule
\end{tabular}
\label{tab:correlation}
\end{table}
\normalsize

These findings provide insight on aspects of the \textit{ARWFML} method that could benefit from more detailed explanations or improved educational resources. The perceived level of difficulty or ease by the participants in responding to the questions highlights the areas requiring improvement. In summary, the language concepts and models created with \textit{ARWFML} are highly comprehensible. For model understanding, responses were accurate in more than 80\% of instances, with an average correctness exceeding 80\% across eight of ten domains, and surpassing 90\% in four domains regarding general language comprehension. There were no discernible differences in the number of correct answers considering different levels of familiarity with augmented reality or conceptual modeling. This suggests that deep knowledge in these areas is not necessary for understanding \textit{ARWFML} models created on the basis of the improved 3D notation. 

\textbf{Summary:} In summary, the newly introduced 3D notation of \textit{ARWFML} and its associated language concepts are in general highly comprehensible for users with various levels of knowledge in augmented reality and conceptual modeling.


\textbf{Threats to Validity}: To assess the comprehensibility of \textit{ARWFML}, a controlled experiment was conducted with the objective of ensuring both internal and external validity. In order to ensure external validity, participants with at least minimal modeling knowledge were recruited. However, the university course setting and predefined use cases may limit the generalizability of the results to real-world contexts. Moreover, the evaluation of the methodology through the use of screenshots rather than hands-on interaction with \textit{M2AR} may result in a limitation of the level of comprehension that could be obtained through active engagement. Furthermore, the emphasis on simple scenarios raises concerns about the applicability of the methodology to complex situations.

With regard to internal validity, the use of screenshots to test model understanding may not fully measure \textit{ARWFML}'s performance. Direct interaction with models could provide authenticity, but also introduce biases in terms of comprehensibility. Furthermore, participant selection may introduce bias due to varying proficiency levels. Finally, sequencing the prototype introduction before administering the questionnaire could influence perceptions, but the interdependence between the language and the application may mitigate this.

\section{Conclusion and Outlook}
\label{sec:conclusion}

In this paper, we explore how modeling languages for the domain of augmented reality can be evaluated from multiple perspectives. By using the example of the recently developed \textit{ARWFML} language for modeling AR workflows, we could illustrate how such an evaluation can be conducted in practice. One of the key insights that can be derived is that the empirical evaluation of such languages requires the availability of adequate tool support. This principle also applies to the assessment of a language's comprehensibility. The reason is that modeling languages for augmented reality typically require the specification of spatial information in three-dimensional space, which cannot be accomplished in 2D modeling environments. The introduction of a new 3D-enabled modeling platform allows for the creation of formally correct models that were subsequently utilized as input for an empirical user study. The study showed that the concepts of the \textit{ARWFML} language are well comprehensible even by non-experts in the domain of augmented reality.

The next step in a subsequent design cycle will be the extension of the \textit{ARWFML} language towards supporting further workflow patterns. Furthermore, we plan to advance the usability of the \textit{M2AR} modeling platform so that further empirical studies can be conducted with it. This will permit, for example, to assess whether the interaction with \textit{ARWFML} models helps to even better understand the working of such a language.

\newpage
%
%
%
\bibliographystyle{splncs04}
\bibliography{bibliography}

\end{document}